\pdfoutput=1

\documentclass[11pt]{article}

\usepackage[preprint]{acl}

\usepackage{times}
\usepackage{latexsym}

\usepackage[T1]{fontenc}

\usepackage[utf8]{inputenc}

\usepackage{microtype}

\usepackage{inconsolata}

\usepackage{graphicx}

\title{GIT-CXR: End-to-End Transformer \\
for Chest X-Ray Report Generation}

\author{
 \textbf{Iustin Sîrbu\thanks{Equal contribution.}\textsuperscript{1}}$\mbox{  }$
 \textbf{Iulia-Renata Sîrbu\footnotemark[1]\textsuperscript{1,2}}$\mbox{  }$
 \textbf{Jasmina Bogojeska\textsuperscript{2}}$\mbox{  }$
 \textbf{Traian Rebedea\textsuperscript{1}}
\\
\\
 \textsuperscript{1}National University of Science and Technology POLITEHNICA Bucharest, Romania
 \\
 \textsuperscript{2}ZHAW School of Engineering, Winterthur,
Switzerland
\\
 \small{
   \textbf{Correspondence:} \href{mailto:email@domain}{iustin.sirbu@upb.ro}
 }
}

\usepackage{booktabs}
\usepackage{graphicx} 
\usepackage{subcaption} 
\usepackage{multirow}
\usepackage{tabularray}

\begin{document}
\maketitle
\begin{abstract}
Medical imaging is crucial for diagnosing, monitoring, and treating medical conditions. The medical reports of radiology images are the primary medium through which medical professionals attest their findings, but their writing is time consuming and requires specialized clinical expertise. The automated generation of radiography reports has thus the potential to improve and standardize patient care and significantly reduce clinicians workload. Through our work, we have designed and evaluated an end-to-end transformer-based method to generate accurate and factually complete radiology reports for X-ray images. Additionally, we are the first to introduce curriculum learning for end-to-end transformers in medical imaging and demonstrate its impact in obtaining improved performance. The experiments have been conducted using the MIMIC-CXR-JPG database, the largest available chest X-ray dataset. The results obtained are comparable with the current state-of-the-art on the natural language generation (NLG) metrics BLEU and ROUGE-L, while setting new state-of-the-art results on F1 examples-averaged, F1-macro and F1-micro metrics for clinical accuracy and on the METEOR metric widely used for NLG.
\end{abstract}

\section{Introduction}

Interpreting radiography images with complex and detailed features is a challenging task that demands significant time \cite{jing2017automatic, li2018hybrid} and specialized clinical expertise \cite{delrue2011difficulties}. The insights clinicians provide in these reports are crucial for future patient assessments, and errors can result in misdiagnoses and improper treatment. Consequently, the increasing volume of radiographic images, particularly in public hospitals and densely populated areas, coupled with a limited number of experts, leads to substantial delays, negatively affecting diagnosis and treatment outcomes.

Radiology report generation can be framed as an image captioning task. While recent advances in image captioning, such as CNN encoders paired with RNN decoders \cite{jing2017automatic, li2018hybrid} and transformers integrated with complex modules \cite{cao2023mmtn}, have shown promise, generating accurate medical reports from images remains an unresolved challenge, with current performance levels insufficient for practical use. The main challenges in medical image captioning include highly similar complex images with subtle differences, the use of domain-specific language, and brief diagnostic insights embedded within lengthy repetitive descriptions. 

This work focuses on developing and evaluating an end-to-end transformer approach, designed specifically for the automated generation of radiology reports from radiography images.
We adapt the GIT transformer \cite{wang2022git} by incorporating widely-used techniques such as adding a classification head, using patient's history and multi-view images. Most distinctively, we integrate curriculum learning in our training and prove it's efficacy. Together with these enhancements, we obtain an end-to-end transformer approach that outperforms existing methods, while experimentally confirming that all these ingredients are essential for our method's success.
We emphasize that the techniques used by us don't add extra training or inference complexity to the network, as opposed to existing approaches \cite{tanida2023interactive,bu2024instance}.  

Current research encounters great problems with the generation of long medical reports \cite{zhao2023radiology}. We substantiate the essential role curriculum learning plays in providing significant improvement for this obstacle and consider that this specific issue was not properly addressed by previous works and it should be a topic of great interest for the medical community.

 We conduct our experiments on the largest available chest X-ray dataset - the MIMIC-CXR-JPG dataset introduced by \citet{johnson2019mimic}. The results obtained with our proposed solution have set a new state-of-the-art on the natural language generation (NLG) metric METEOR \cite{banerjee2005meteor}, 
and on the clinical accuracy metrics F1-macro and F1-micro. 
This demonstrates both the clinical accuracy and factual completeness of our generated reports. Furthermore, our results are on par with the state-of-the-art considering the natural language generation metrics BLEU \cite{papineni2002bleu} and ROUGE-L \cite{lin2004rouge}.

Our main contributions can be summarised as follows: 
    \begin{itemize}
        \item We propose an \textbf{end-to-end transformer} approach for the generation of medical reports for chest X-ray images, proving the validity of simpler architectures.
        \item To the best of our knowledge, we are the first to show the effectiveness of \textbf{curriculum learning} for the task of automated radiology report generation using transformers.
        \item We show the capacities of our setups by obtaining \textbf{state-of-the-art results}, over the largest benchmark of chest radiography, MIMIC-CXR-JPG, for both clinical accuracy metrics as well as natural language generation metrics, attesting both the factual completeness as well as the accuracy of our generated reports.
     \end{itemize}
    
\begin{figure*}[t]
    \vspace{-2mm}
     \centering
     \begin{subfigure}{0.65\textwidth}
         \centering
         \includegraphics[width=\textwidth, height=\textheight, keepaspectratio]{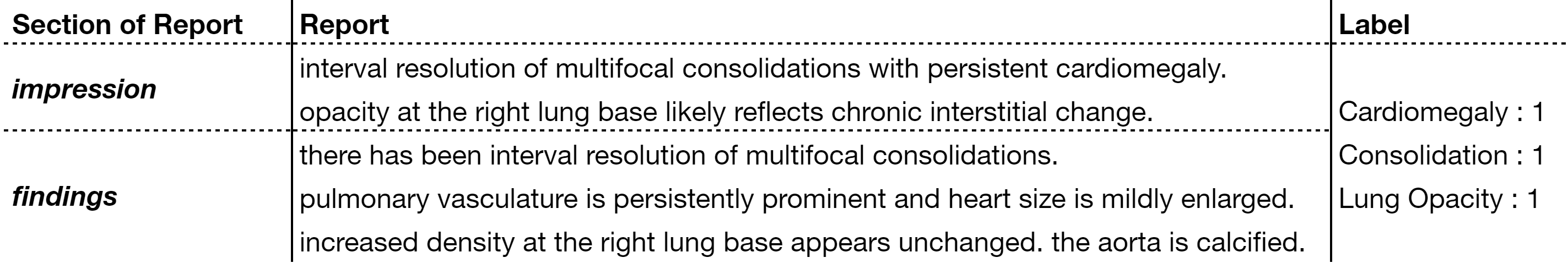}
         \caption{Example of a medical report with text pre-processing and corresponding labels.}
         \label{fig:gen_report_ex}
     \end{subfigure}
     \hfill
     \begin{subfigure}{0.11\textwidth}
         \centering
         \includegraphics[width=\textwidth, height=\textheight, keepaspectratio]{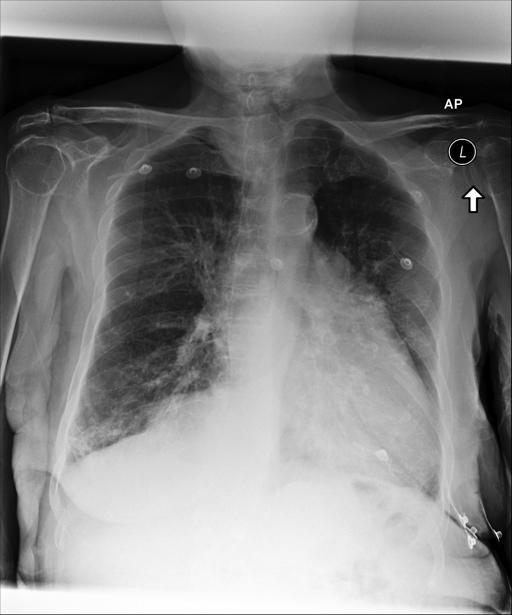}
         \caption{AP}
         \label{fig:ap}
     \end{subfigure}
     \begin{subfigure}{0.11\textwidth}
         \centering
         \includegraphics[width=\textwidth, height=\textheight, keepaspectratio]{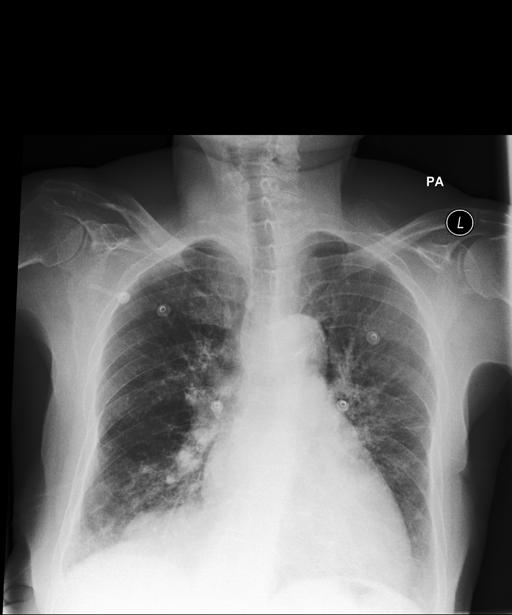}
         \caption{PA}
         \label{fig:pa}
     \end{subfigure}
       \begin{subfigure}{0.11\textwidth}
         \centering
         \includegraphics[width=\textwidth, height=\textheight, keepaspectratio]{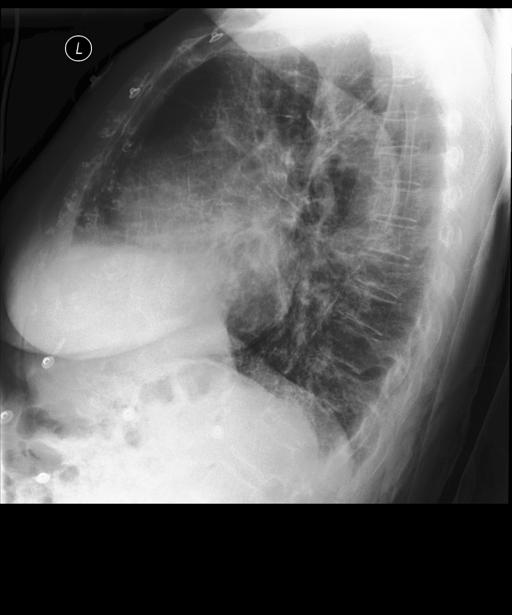}
         \caption{LAT}
         \label{fig:lat}
     \end{subfigure}
     \caption{Example of a study of one patient.}
     \label{fig:study}
\end{figure*}

\section{Related Work}
\label{sec:related_work}

\subsection{Transformers in Image Captioning.} Attention models have gained large-scale popularity for the task of image captioning \cite{cornia2020meshed,nguyen2022grit,zhang2021rstnet}, due to their outstanding performance. The GIT Transformer \cite{wang2022git} is a generative image-to-text transformer that has obtained state-of-the-art results on various computer vision tasks. It has been pre-trained on 0.8B image-text pairs from various sources, but as far as we know has not been tested on medical imaging tasks.
The architecture of GIT is based on two transformer modules - an image encoder - based on the vision transformer model of \citet{yuan2021florence} - that extracts the features of the input image, and a text decoder (also a transformer module) which uses the image features, in order to generate the image caption.

\subsection{Radiology Report Generation} 

Radiology report generation from radiography images stands in the broader task of image captioning \cite{vinyals2015show, xu2015show}. The majority of methods used to solve the challenges of medical image captioning are based on deep learning models that use an image-encoder and text-decoder architecture. The encoder is most often a Convolutional Neural Network (CNN) \cite{li2019knowledge, li2018hybrid, srinivasan2020hierarchical, yin2019automatic}, used to encode the features of the images and create their latent representations. The decoder - a Recurrent Neural Network (RNN), such as LSTM \cite{hochreiter1997long} - is then used to convert the extracted features into the generated reports \cite{jing2017automatic}. 

Transformers \cite{vaswani2017attention} bring the advantage of effectively learning long-range visual and textual dependencies and are successfully used for this task, but models using them often employ additional CNN encoders \cite{miura2020improving, lovelace2020learning}, or complex modules (e.g. added memory augmentation module \cite{cao2023mmtn}, Faster R-CNN object detector \cite{tanida2023interactive}, interpreter-generator-classifier modules \cite{nguyen2021automated}, relational memory \cite{chen2020generating}), that add inference complexity and require more supervision during training. 

Therefore, to address these problems that come with the added complexity, we decided to research the performance of a simpler approach, an end-to-end transformer. In the context of medical image captioning, this approach has not been studied in depth until recently by \citet{wang2022automated} and \citet{nicolson2023improving}, but such works do not address the issue of short generated reports. 

\subsection{Curriculum Learning} 
First introduced by \citet{bengio2009curriculum}, the vanilla form of curriculum learning means to gradually increase the difficulty of the data samples that the model sees during training. This is similar to the form in which humans initially acquire knowledge. 

The effectiveness of curriculum learning for text generation has been previously studied by \citet{subramanian2017adversarial}. They show the importance of constraining their adversarial model to generate increasingly longer sequences. The sequence length has also been used as a difficulty metric for curriculum learning in other NLP tasks. \citet{spitkovsky2009baby} shows the importance of an easy-to-hard training strategy for unsupervised grammar induction, while \citet{chang2021does} applies curriculum learning for data-to-text generation and shows that it improves both generation quality and convergence speed.

In the field of medical imaging, curriculum learning has been mostly used on computer vision tasks, by employing handcrafted curriculum, or an order based on human annotators \cite{lotter2017multi, jimenez2022curriculum, oksuz2019automatic, wei2021learn}. More recently, \citet{alsharid2020curriculum} employed a dual-curriculum approach for the task of fetal ultrasound image captioning, using the Wasserstein distance for image data and the TF-IDF metric \cite{sparck1972statistical} for text data. \citet{liu2022competence} applied curriculum learning for generating medical reports, by employing an iterative 2-steps approach: first, estimate the difficulty of the training samples and evaluate the competence of the model; second, select the appropriate training samples by following the easy-to-hard strategy. 

However, curriculum learning has not been studied previously in the context of medical reports generation using transformers. Inspired by the curriculum methods used in NLP and by the observation that longer reports are harder to generate for our baseline models, we introduce a curriculum learning approach based on report length and show that it is as effective as it is simple, as opposed to the multimodal curriculum usually employed for medical imaging.

\section{Method}
\label{sec:method}

In order to adapt the GIT transformer for the complex task of automated report generation for radiography images, we employ a variety of task-specific methods. While some of them are widely used in the field of medical image captioning (e.g. we combine the approach of using multi-view images with the approach of using the patient's medical history and the approach of training the model in a multi-task setting), we are the first to experiment with a curriculum learning method, based on the report length. All of these methods are further elaborated in this section.

\subsection{GIT-CXR (SV)} The \textbf{single-view} approach is depicted in Figure \ref{fig:cls} when ignoring the Multi-label Classifier. 
This is our baseline method, consisting of fine-tuning  the GIT model for image captioning on the MIMIC-CXR dataset.
Our single-view chest X-ray image is passed through the image encoder obtaining an embedding that is fed to the text decoder to generate a radiology report. In this case, we keep only the AP and PA images and duplicate the corresponding report for each. 


\subsection{GIT-CXR (MV)} For the \textbf{multi-view} approach, we use samples of two images at once, correlated with their corresponding report, meaning combinations of AP, PA, LATERAL and LL images. We enforce that at least one of the images is AP/PA and we duplicate the image of a report, if only one is available.


As shown in Figure \ref{fig:mv}, each single-view image is passed through the image encoder, obtaining an image embedding for each view. A different temporal embedding is added to each view to differentiate the different views of the multi-view image. Finally, the resulting embeddings are concatenated creating a final image embedding for the entire multi-view image. The next steps are similar to the single-view approach described before.

\subsection{GIT-CXR-CLS} 
\begin{figure*}[h!]
\vspace{-2mm}
     \centering 
     \begin{subfigure}{0.5\textwidth}
         \centering
         \includegraphics[width=\textwidth, height=\textheight, keepaspectratio]{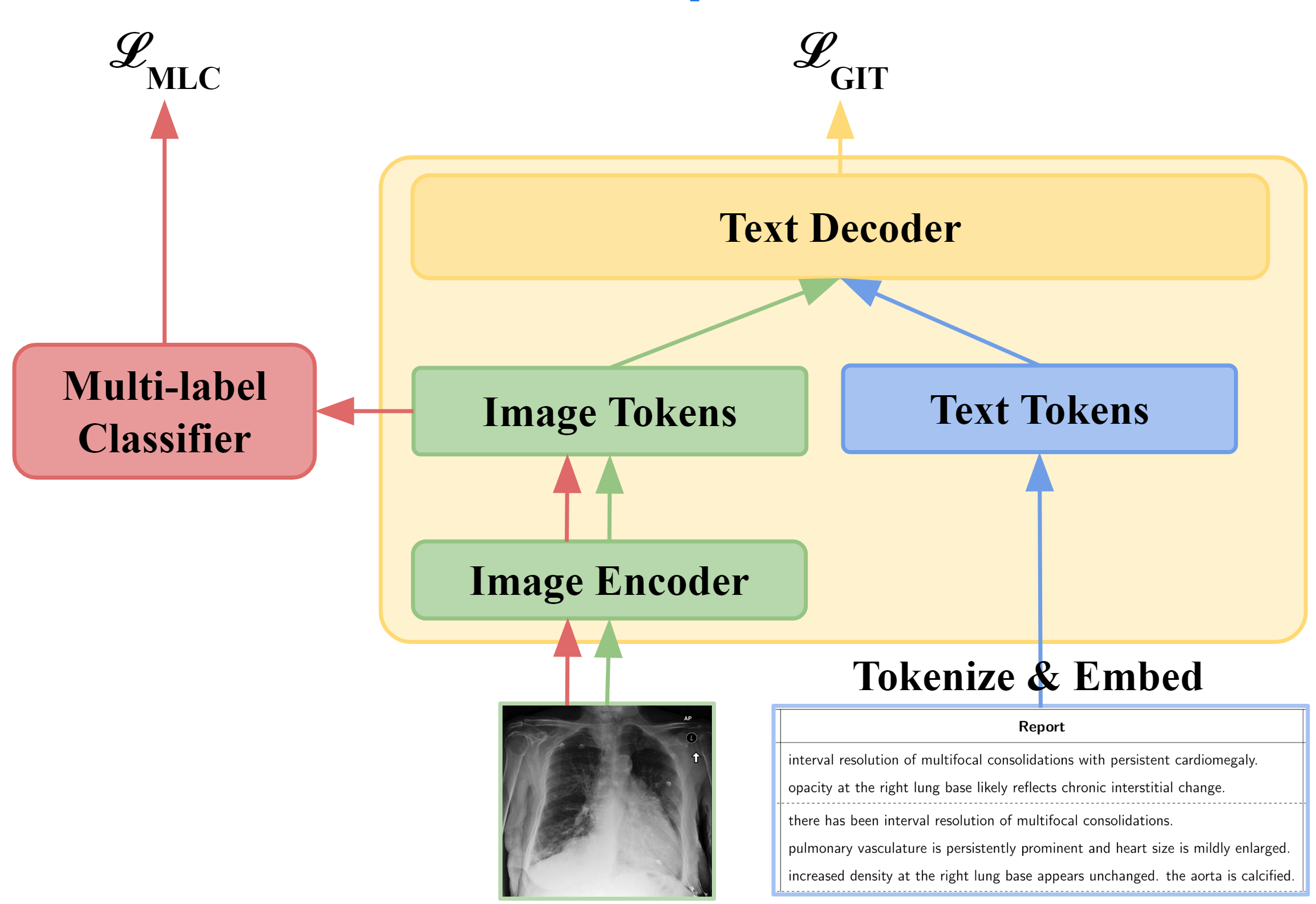}
         \caption{GIT-CXR(SV) and GIT-CXR-CLS(SV) architectures.}
         \label{fig:cls}
     \end{subfigure}
     \hspace{7mm}
     \begin{subfigure}{0.42\textwidth}
         \centering
         \includegraphics[width=\textwidth, height=\textheight, keepaspectratio]{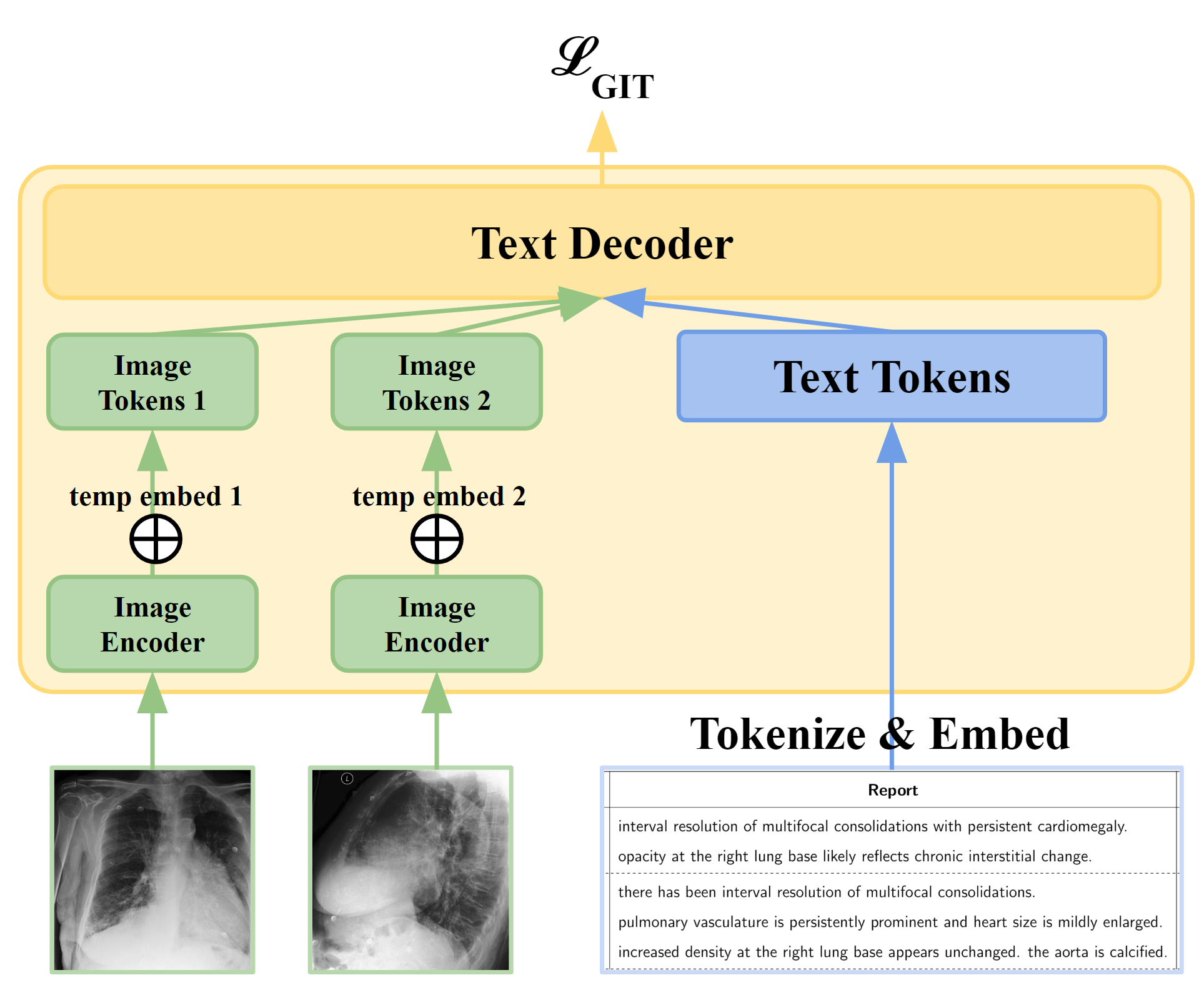}
         \caption{GIT-CXR(MV) architecture with 2 images.}
         \label{fig:mv}
     \end{subfigure}
     \caption{Our proposed methods with their setups.}
     \label{fig:arch}
\end{figure*}

We also introduce an \textbf{auxiliary loss for multi-label classification} which can be seen in Figure \ref{fig:cls}. Similar to \citet{nguyen2021automated}, the output of the image encoder is passed to a multi-label classifier which consists of a classification head for each of the 14 possible diagnostics - given by the CheXbert labeler~\citet{smit2020CheXbert}. We define the classification loss as the mean of the losses computed for each head via weighted Cross Entropy, as shown in Equation \ref{eq:mlc_loss}, where $D$ is the number of pathologies (in our case 14); $x$ is the input image; $f_v(x)$ is the image encoding, given by passing $x_i$ through the vision encoder $f_v$; $h_i(f_v(x))$ is the prediction of the $i$-th classification head; $y_i$ is the target for the diagnostic $i$ and $CE$ is the Cross Entropy loss, where the class weights are computed for each pathology individually.

\vspace{-7mm}

\begin{equation}
    \mathcal{L}_{MLC} = \frac{1}{D} \sum_{i=1}^{i\leq D} CE(y_i, h_i(f_v(x)))
    \label{eq:mlc_loss}
\end{equation} 

\subsection{Context} To all the methods detailed above, we also add context. The context is given together with the target report to be tokenized and fed to the text decoder. The context is obtained by concatenating the \textit{'indication'} and \textit{'history'} fields of the reports. 


\subsection{Curriculum Learning} One of the main challenges of medical report generation is training the model to generate reports that are long enough and include all the relevant information. Indeed, as it can be seen in Figure \ref{generatedtarget}, the generated reports tend to be shorter than the corresponding targets. This leads to a steep decrease in performance when long reports are expected (Figure \ref{fig:metrics_per_length}), which drastically affects the overall performance of the model. 
To address this, we propose a curriculum learning approach based on the length of the target report: we train the model using shorter (and easier to learn) samples in the initial phases and increase the mean length of the target reports from one epoch to another. 

More precisely, we split the dataset in $b$ bins of equal size. During each training epoch, we assign a weight $\frac{1}{1 + |i - i_e|}$ to the samples corresponding to each bin $1 \leq i \leq b$, where $i_e$ is the bin corresponding to the current epoch. Then, we sample without replacement a fraction $f$ of the dataset, using these weights. This ensures that during each epoch, the most samples used will come from the proximity of bin $i_e$, while still allowing for some amount of samples of opposite difficulty level. For example, in the early stages of the training, the model will still be able to see words that might appear only in long reports, while in the final stages of the training the model will still see some easy examples that require the generation of short reports, so it won't overfit to long sentences.

\begin{table*}[t]
\centering
\small
\begin{tabular}{c|cccccccccc}
\toprule
\textsc{Model} & \textsc{BL}\textsubscript{1} & \textsc{BL}\textsubscript{2} & \textsc{BL}\textsubscript{3} & \textsc{BL}\textsubscript{4} & \textsc{RG}\textsubscript{L} & \textsc{M} & \textsc{F1}\textsubscript{MA} & \textsc{F1}\textsubscript{MI} & \textsc{F1}\textsubscript{MI5} & \textsc{F1}\textsubscript{EX}\\
\midrule
\textsc{GIT-CXR (MV+C+CL)} & $0.403$ & \underline{0.286} & \underline{0.215} &\underline{0.168} & $0.312$ & \textbf{0.369} & \textbf{0.348} & \textbf{0.534} & \underline{0.565} & \textbf{0.458} \\
\textsc{GIT-CXR (SV+C+CL)} & $0.393$ & $0.278$ & $0.208$ & $0.162$ & $0.305$ & \underline{0.359} & \underline{0.327} & \underline{0.505} & $0.538$ & $0.428$ \\
\midrule
\textsc{ARR TR \citeyearpar{wang2022automated}} & $0.351$ & $0.223$ & $0.157$ & $0.118$ & $0.287$ & $-$ & $-$ & $-$ & $-$ & $-$ \\
\textsc{{RGRG \citeyearpar{tanida2023interactive}}} & $0.373$ & $0.249$ & $0.175$ & $0.126$ & $0.264$ & $0.168$ & $-$ & $-$ & $0.547$ & \underline{0.447}\\
\textsc{{EKAGen \citeyearpar{bu2024instance}}} & \underline{0.419} & $0.258$ & $0.170$ & $0.119$ & $0.287$ & $0.157$ & $-$ & $0.499$ & $-$ & $-$\\
\textsc{{CvT-212Distil \citeyearpar{nicolson2023improving}}} & $0.392$ & $0.245$ & $0.169$ & $0.124$ & $0.285$ & $0.153$ & $-$ & $-$ & $-$ & $0.384$\\
\textsc{R2GEN \citeyearpar{chen2020generating}} & $0.353$ & $0.218$ & $0.145$ & $0.103$ & $0.277$ & $0.142$ & $0.276$ & $-$ & $-$ & $-$\\
\midrule
\textsc{AGA (MV+T+I) \citeyearpar{nguyen2021automated} $\dagger$} & \textbf{0.495} & \textbf{0.360} & \textbf{0.278} & \textbf{0.224} & \textbf{0.390} & $0.222$ & $-$ & $-$ & $-$ & $-$\\
\textsc{LOVE \citeyearpar{lovelace2020learning} $\dagger$} & $0.415$ & $0.272$ & $0.193$ & $0.146$ & \underline{0.318} & $0.159$ & $0.228$ & $0.411$ & $-$ & $-$\\
\textsc{MMTN \citeyearpar{cao2023mmtn} $\dagger$} & $0.379$ & $0.238$ & $0.159$ & $0.116$ & $0.283$ & $0.161$ & $-$ & $-$ & $-$ & $-$\\
\textsc{CXR-RePaiR \citeyearpar{endo2021retrieval} $\dagger$} & $-$ & $0.069$ & $-$ & $-$ & $-$ & $-$ & $0.274$ & $-$ & $-$ & $-$\\
\textsc{$M2$TR \citeyearpar{miura2020improving} $\dagger$} & $-$ & $-$ & $-$ & $0.133$ & $-$ & $-$ & $-$ & $-$ & \textbf{0.567} & $-$\\
\bottomrule
\end{tabular}
\caption{Results on the full MIMIC-CXR-JPG dataset \cite{johnson2019mimic}. The models with '$\dagger$' \textbf{don't} use the original splits. The best results for each task are highlighted using {\bf bold} font and the second best with \underline{underline}. Apart from us, the ARR TR approach is the only end-to-end transformer architecture. All of the architectures below it use additional CNN/LSTM \cite{hochreiter1997long} modules. If not specified otherwise, we report all the results from the original papers. All of our results are the average of three training runs.} 
\label{results}
\end{table*}

\begin{table*}[t]
\vspace{-1mm}
\centering
\small
\begin{tabular}{c|cccccccccc}
\toprule
\textsc{Model} & \textsc{BL}\textsubscript{1} & \textsc{BL}\textsubscript{2} & \textsc{BL}\textsubscript{3} & \textsc{BL}\textsubscript{4} & \textsc{RG}\textsubscript{L} & \textsc{M} & \textsc{F1}\textsubscript{MA} & \textsc{F1}\textsubscript{MI} & \textsc{F1}\textsubscript{MI5} & \textsc{F1}\textsubscript{EX}\\
\midrule
\textsc{GIT-CXR-CLS (MV+C+CL)} & $0.389$ & $0.274$ & $0.205$ & $0.159$ & $0.302$ & $0.357$ & $0.318$ & $0.495$ & $0.530$ & $0.420$\\
\textsc{GIT-CXR-CLS (SV+C+CL)} & $0.386$ & $0.273$ & $0.204$ & $0.159$ & $0.301$ & $0.355$ & $0.312$ & $0.486$ & $0.513$ & $0.411$\\
\textsc{GIT-CXR (MV+C+CL)} & \textbf{0.403} & \textbf{0.286} & \textbf{0.215} & \textbf{0.168} & \textbf{0.312} & \textbf{0.369} & \textbf{0.348} & \textbf{0.534} & \textbf{0.565} & \textbf{0.458}\\
\textsc{GIT-CXR (SV+C+CL)} & \underline{0.393} & \underline{0.278} & \underline{0.208} & \underline{0.162} & $0.305$ & \underline{0.359} & \underline{0.327} & \underline{0.505} & \underline{0.538} & \underline{0.428}\\
\textsc{GIT-CXR-CLS (MV+C)} & $0.354$ & $0.254$ & $0.193$ & $0.152$ & $0.310$ & $0.351$ & $0.308$ & $0.486$ & $0.526$ & $0.410$\\
\textsc{GIT-CXR-CLS (SV+C)} & $0.352$ & $0.252$ & $0.189$ & $0.149$ & $0.307$ & $0.348$ & $0.313$ & $0.487$ & $0.516$ & $0.412$\\
\textsc{GIT-CXR (MV+C)} & $0.343$ & $0.248$ & $0.188$ & $0.149$ & \underline{0.311} & $0.347$ & $0.298$ & $0.462$ & $0.496$ & $0.386$\\
\textsc{GIT-CXR (SV+C)} & $0.324$ & $0.230$ & $0.172$ & $0.136$ & $0.294$ & $0.331$ & $0.257$ & $0.407$ & $0.428$ & $0.334$\\
\textsc{GIT-CXR (MV)} & $0.316$ & $0.199$ & $0.130$ & $0.090$ & $0.240$ & $0.291$ & $0.294$ & $0.495$ & $0.536$ & $0.415$\\
\textsc{GIT-CXR (SV)} & $0.299$ & $0.187$ & $0.122$ & $0.084$ & $0.235$ & $0.282$ & $0.262$ & $0.452$ & $0.500$ & $0.376$\\
\bottomrule
\end{tabular}
\caption{Results of the ablation study on the full MIMIC-CXR-JPG dataset \cite{johnson2019mimic} using the original splits. The best results for each task are highlighted using {\bf bold} font and the second best with \underline{underline}.}
\label{resultsablation}
\end{table*}

\section{Experiments}

\subsection{Dataset} 
\label{sec:dataset}

The MIMIC Chest X-ray Database v2.0.0 \cite{johnson2019mimicorig}, is the largest publicly available dataset containing chest X-ray images with their corresponding, free-text, clinical reports. The dataset contains a total of 377,110 DICOM format radiography images that correspond to 227,835 studies of 64,588 patients. This dataset is the basis of the MIMIC-CXR-JPG dataset, or MIMIC Chest X-ray JPG Database v2.0.0 \cite{johnson2019mimic}, that is entirely derived from MIMIC-CXR. Additionally processed, it provides JPG conversion of the original DICOM images and 14 pathologies (labels) for the reports using the CheXpert labeler~ \cite{irvin2019chexpert}. All of the 14 pathologies have 4 possible classes each (\textit{'Positive'}, \textit{'Negative'}, \textit{'Uncertain'} and \textit{'Missing'}). The MIMIC-CXR-JPG dataset offers the standard reference splits we also adopt in our study.

As previous works use different sections of the reports (e.g. \citet{endo2021retrieval} uses both the \textit{'findings'} and the \textit{'impression'}, whereas \citet{nguyen2021automated}, \citet{lovelace2020learning} and \citet{miura2020improving} use only the \textit{'findings'} field), we decided to also use the most complete information available, so we concatenate the 'impression' and 'findings' sections and drop the studies that don't contain any of them. An example of a study comprising 3 different images, a report with 'impression' and 'findings' sections and their associated CheXpert labels can be seen in Figure \ref{fig:study}.
Additional information about the dataset and about our data processing can be found in Appendix \ref{appendix:dataset}.

\subsection{Evaluation Metrics} 
\label{sec:evaluation_metrics}

NLG metrics are the default evaluation approach used for image captioning tasks in order to evaluate the ability of the model to generate coherent text. While BLEU \cite{papineni2002bleu} is precision-based and ROUGE \cite{lin2004rouge} is recall-based, METEOR \cite{banerjee2005meteor} provides a way of combining both precision and recall and it is also better correlated with human judgement. However, as \cite{boag2020baselines} found that it is possible to have models with high NLG scores that don't produce correct diagnosis, clinical accuracy metrics have been introduced in order to measure the ability of a model to produce reports that could be used to identify the right pathologies. Therefore, for the most complete comparison to previous works and in order to determine both the factual completeness as well as the clinical accuracy of our approach, the metrics we decided to use are both NLG metrics - BLEU, ROUGE-L and METEOR as well as the clinical accuracy metrics F1-macro and F1-micro on all 14 labels, on just the 5 most frequent labels (to compare  with \cite{miura2020improving}) and finally the F1 examples-averaged. 

Because our F1 score needs to be computed in a multi-label (the 14 labels of CheXbert) and multi-class (\textit{Positive}, \textit{Negative}, \textit{Uncertain} and \textit{Missing}) manner, we proceed similar to \cite{lovelace2020learning}: we compute the F1 score of the \textit{Positive} class for each pathology individually and then we compute their macro-average - for the comparison with \cite{endo2021retrieval} and \cite{lovelace2020learning} - and micro-average for the comparison with \cite{lovelace2020learning}. Although \cite{chen2020generating} reports an F1 score, it is not mentioned weather it is a macro or micro score, while \cite{miura2020improving} only reports the F1-micro on the 5 most frequent labels of the CheXbert labeler. 
More recently, \citet{tanida2023interactive} and \citet{nicolson2023improving} used the example-based average for the reported results. In order to ensure comparison with all methods, we report all the aforementioned F1 averages: F1\textsubscript{MA}, F1\textsubscript{MI}, F1\textsubscript{MI5} and F1\textsubscript{EX}. 


\subsection{Experimental Setup} We train our models on an NVIDIA H100 GPU with 80GB VRAM. For all the experiments, the images were resized to $224\times224$ and the target reports were truncated to 192 tokens.
The full details of the experimental setup are described in Appendix \ref{appendix:experimental_setup}.
We will make our code publicly available. 

\section{Results and Discussion} 

We compare our two best performing methods (GIT-CXR (MV+C+CL) and GIT-CXR (SV+C+CL)) with ten current state-of-the-art works for the task of medical report generation from X-ray images, that also use the MIMIC-CXR-JPG dataset. The results can be seen in Table \ref{results}. We obtain state-of-the-art results with our method GIT-CXR (MV+C+CL), on the NLG metric METEOR, surpassing AGA \cite{nguyen2021automated} by 14.7 percentage points (pp). Moreover, we also exceed all the other methods that reported the clinical accuracy scores F1 examples-averaged (F1\textsubscript{EX}), F1-macro (F1\textsubscript{MA}) and F1-micro (F1\textsubscript{MI}) for the 14 labels. We are only 0.2 pp behind M$2$TR in terms of F1\textsubscript{MI} on the 5 most frequent labels, keeping in mind that they do not use the official splits of the dataset. Between the pure transformer-based approaches, our method obtains state-of-the-art performance on \textbf{all} the metrics presented, outperforming ARR TR \cite{wang2022automated} by at least 5.0pp in terms of BLEU scores and 2.5pp in terms of ROUGE-L. 


\begin{figure*}[h!]
     \centering
     \begin{subfigure}{0.30\textwidth}
         \centering
         \includegraphics[width=\textwidth, height=\textheight, keepaspectratio]{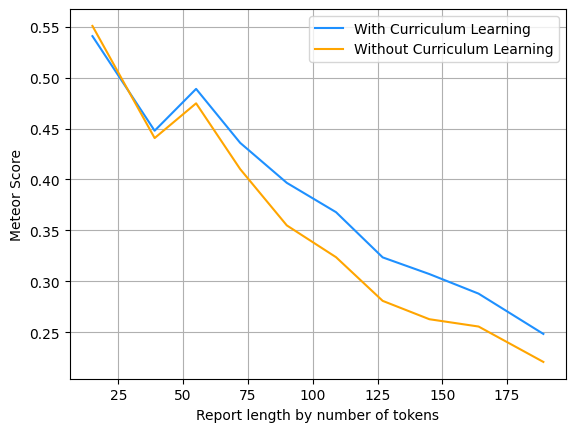}
         \caption{METEOR}
         \label{fig:METEOR_per_length}
     \end{subfigure}
     \begin{subfigure}{0.30\textwidth}
         \centering
         \includegraphics[width=\textwidth, height=\textheight, keepaspectratio]{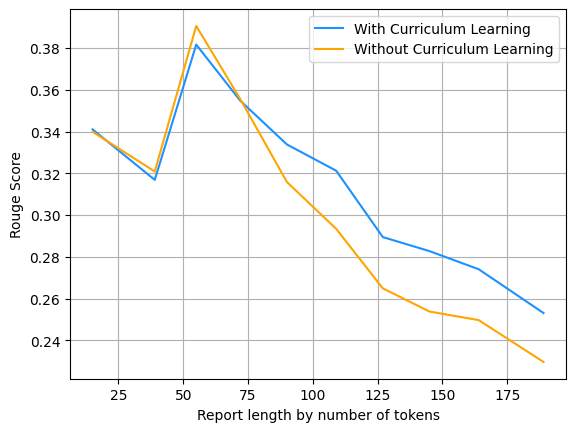}
         \caption{ROUGE-L}
         \label{fig:rouge_per_length}
     \end{subfigure}
       \begin{subfigure}{0.30\textwidth}
         \centering
         \includegraphics[width=\textwidth, height=\textheight, keepaspectratio]{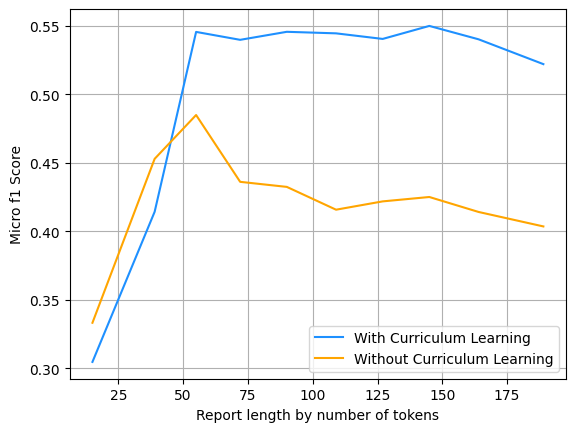}
         \caption{Micro F1}
         \label{fig:microf1_per_length}
     \end{subfigure}
     \caption{The metrics score evolution with report length.}
     \label{fig:metrics_per_length}
\end{figure*}

\subsection{Ablation Study} 
Through our ablation study in Table \ref{resultsablation}, we have proved that all 4 techniques that we introduced (adding context, using multi-view images, adding a multi-label classifier and using curriculum learning) have a positive impact on the performance, both in terms of the NLG metrics and the clinical accuracy metrics. Compared to the baseline (GIT-CXR (SV)), adding the context improves all the scores and methods, adding a classification head improves the clinical accuracy metrics the most, using the multi-view method improves the results of the single-view method and using curriculum learning improves all of these methods the most.

\paragraph{Curriculum Learning Impact} An important finding in our ablation study is that while curriculum learning does improve the performance of our approach, the curriculum learning architectures with the added multi-label classifier (GIT-CXR-CLS (SV+C+CL) and GIT-CXR-CLS (MV+C+CL)) perform worse than the curriculum learning architectures without the multi-label classifier. This odd occurrence breaks the pattern we have seen so far that the added classification head is an overall improvement over all of our proposed methods. Even though we thought it a given that GIT-CXR-CLS (MV+C+CL) would be the best performing architecture as it respected all the improvement patterns before it, we observe experimentally that the classification head is not compatible with the curriculum learning method we employed. We address this on the fact that our curriculum methodology radically changes the pathologies distribution seen by the classification head during each epoch. For medical reports the problem is that short reports are correspondent to healthy patients that didn't need detailed explanations in their findings. However, in the case of ill patients with many underlying conditions, the reports are inevitably longer. While the train distribution is already very skewed, with half of the pathologies being present in less than 5\% of the patients, the curriculum learning based solely on the text length provides the multi-label classifier with a more extreme version of it, in the early stages of the training. This is a very interesting discovery that highlights even more the particularities of medical text as opposed to a more general field. 

\begin{figure}[h!]
     \centering
     \begin{minipage}[b]{0.30\textwidth}
         \centering
         \includegraphics[width=\textwidth, height=0.5\textheight, keepaspectratio]{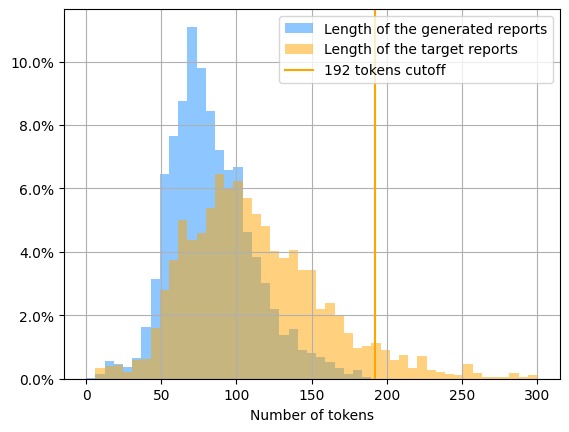}
     \end{minipage}
     \caption{Distributions for the length of the generated reports (blue) and the length of the target reports (yellow), which have been truncated to 192 tokens (orange vertical line).}
     \label{generatedtarget}
\end{figure}

We also test our hypothesis that curriculum learning leads to better performance on generating long reports and make a direct comparison between \textsc{GIT-CXR (MV+C)} - trained without curriculum learning - and \textsc{GIT-CXR (MV+C+CL)} - the corresponding model trained with curriculum learning. In Figure \ref{fig:metrics_per_length}, we show how the evaluation metrics change together with the length of the target report. First, with regards to METEOR (Figure \ref{fig:METEOR_per_length}) and ROUGE-L (Figure \ref{fig:rouge_per_length}), we can see how the two models perform similarly for reports up to 75 tokens. For longer sequences, while both models have a linear decrease in performance, the model trained with curriculum learning has a less abrupt decline. Second, with regards to F1-micro (Figure \ref{fig:microf1_per_length}), we can see how the two models perform similarly bad for very short sequences, but this is mostly because they are rare and the metrics are not very reliable (Figure \ref{generatedtarget}). From 50 tokens onwards, the model trained without curriculum learning is affected by the increase of the target length, while the model that uses curriculum learning is able to maintain a constantly high performance even for very long sequences.  

To conclude, we prove that the reports generated by our best model, trained with curriculum learning, have a high clinical accuracy and can reliably be used for diagnosis. Moreover, our results show that our newly introduced curriculum learning technique has a greater positive impact than the widely used method of adding a classification head (e.g. GIT-CXR (MV+C+CL) vs. GIT-CXR-CLS (MV+C)). Therefore, curriculum learning is an impactful novel advancement to the X-ray report generation task. A future research path would be searching for ways of mixing the two methods so a model could benefit form both of them simultaneously.

\begin{table*}[h!]
\small
\centering
\begin{tabular}{c|cccccccc}
\toprule
\textsc{Category} & \textsc{F1} & \textsc{P} & \textsc{R} & \textsc{Support}  \\ \hline
        \textsc{Enlarged Cardiomediastinum} & $0.088$ & $0.203$ & $0.057$ & $230$  \\ 
        \textsc{Cardiomegaly} & $0.642$ & $0.627$ & $0.658$ & $1168$  \\ 
        \textsc{Lung Opacity} & $0.478$ & $0.532$ & $0.434$ & $1131$  \\ 
        \textsc{Lung Lesion} & $0.148$ & $0.333$ & $0.096$ & $178$  \\ 
        \textsc{Edema} & $0.461$ & $0.492$ & $0.433$ & $695$  \\ 
        \textsc{Consolidation} & $0.162$ & $0.262$ & $0.118$ & $187$  \\ 
        \textsc{Pneumonia} & $0.259$ & $0.282$ & $0.239$ & $213$  \\ 
        \textsc{Atelectasis} & $0.432$ & $0.463$ & $0.404$ & $890$  \\ 
        \textsc{Pneumothorax} & $0.310$ & $0.310$ & $0.310$ & $71$  \\ 
        \textsc{Pleural Effusion} & $0.669$ & $0.676$ & $0.661$ & $1116$  \\ 
        \textsc{Pleural Other} & $0.113$ & $0.296$ & $0.070$ & $114$  \\ 
        \textsc{Fracture} & $0.033$ & $0.143$ & $0.019$ & $161$  \\ 
        \textsc{Support Devices} & $0.766$ & $0.776$ & $0.757$ & $1327$  \\ 
        \textsc{No Finding} & $0.317$ & $0.244$ & $0.451$ & $193$  \\ \midrule
        \textsc{MACRO\_AVG} & $0.349$ & $0.403$ & $0.336$ &   \\ 
        \textsc{MICRO\_AVG} & $0.537$ & $0.573$ & $0.506$ & ~ \\ 
\bottomrule
\end{tabular}
\caption{Clinical accuracy metrics per pathology for our best model GIT-CXR (MV+C+CL).}
\label{resultsperlabel}
\end{table*}

\vspace{-3mm}

\subsection{Reports Analysis} 

In Table \ref{labelan}, from Appendix \ref{appendix:qualitative_analysis}, we extracted three examples of generated reports with their corresponding context and ground truth report, using the GIT-CXR (MV+C+CL) approach.

The first example is a complete report that contains both the \textit{'impression'} and the \textit{'findings'} sections. The results are overall good for the ROUGE-L (0.349) and METEOR (0.330) but BLEU, however, is much lower compared to our overall method's performance (BLEU-1 is 0.165 compared to 0.396). This is a perfect example of how the length of the target report compared to the generated report affects the BLEU score. As the generated reports are too short, the brevity penalty used by the BLEU score will be low, which has a high impact on all the BLEU scores.

The second example misses the \textit{'impression'} section in the target report, but despite this, our approach is able to generate with descent accuracy the \textit{'findings'} section and leaves the \textit{'impression'} section empty, therefore all of the results are extremely high on all NLG metrics (BLEU-1 - 0.571, ROUGE-L - 0.492, METEOR - 0.701).

The final example contains a very short context \textit{'picc.'} and both the \textit{'impression'} and the \textit{'findings'} sections of the target report. The results in this case are very poor on all of the NLG metrics. That is because the context is not only too small to count but the target report is very large and our approach generated a much shorter report, with an empty \textit{'impression'} section. Additionally, the METEOR score is also very low (0.162), as the overall meaning of the generated report doesn't match the one of the target.

\subsection{Labels Analysis} As there are 13 pathologies that may appear in the reports (14 if we include the 'No Findings' label), it is obvious that a model may have very different performances in identifying each one of them, for example because they are differently represented in the dataset and even because some of them might be easier to identify from an image. Therefore, we conduct an analysis of how well our best model performs for each individual label, based on the results shown in Table \ref{resultsperlabel}.

First, we notice that the precision score tends to be higher, with a minimum value of $0.143$ for the \textsc{Fracture} label, whereas the recall reaches values smaller than $0.1$ on multiple occasions (\textsc{Enlarged Cardiomediastinum}, \textsc{Lung Lesion}, \textsc{Pleural Other} and \textsc{Fracture}). This shows a tendency of the model to fail to identify a pathology rather than wrongly mention it in the report. This is also supported by the fact that the precision is considerably greater than the recall for multiple categories, such as \textsc{Enlarged Cardiomediastinum} (3.5 times larger), \textsc{Lung Lesion} (3.5 times larger), \textsc{Consolidation} (2.2 times larger), \textsc{Pleural Other} (4.2 times larger) or \textsc{Fracture} (7.5 times larger).

Second, we notice that there is also a correlation between the support of a label and the big difference between the corresponding precision and recall. More precisely, for all the labels with a support smaller than 300 samples, with the exception of \textsc{Pneumothorax} and \textsc{No Finding} (which is the absence of pathologies), the precision is considerably larger than the recall. On the other hand, for the better represented categories (over 300 samples), the precision and recall are usually close to one another, with the recall even surpassing the precision in the case of \textsc{Cardiomegaly}. 

Therefore, we conclude that the main reason for the poor performance on some pathologies is their poor representation in the dataset, rather than the inherent difficulty of the task. The value of 300 samples also marks a threshold for about 10\% of the dataset, as this analysis is done on the test set comprising 3082 samples. The poor performance being correlated with the underrepresented labels also justifies the big gap between the macro average (that weights all labels equally) and the considerably larger micro average (that weights all samples equally). This further demonstrates that computing the metrics based solely on the five most represented labels \cite{miura2020improving} yields better scores.

\section{Conclusion}

In this work we have developed and tested a pure transformer-based approach utilising the GIT transformer combined with additional components to address the task of X-ray medical report generation. Curriculum learning was particularly beneficial for the enhanced performance on more difficult samples that necessitated the model to generate a long report. In our experiments, we consider adding many relevant techniques and obtain results that surpass current state-of-the-art methods on application-relevant metrics such as the NLG metric METEOR and clinical accuracy metric F1. We also achieve on par results on other NLG metrics such as BLEU and ROUGE-L. We conduct an extensive ablation study as well as a comprehensive analysis of the labels and generated reports in order to better understand the upsides and downsides of our approach. Our work paves the way for further research and improvements along this direction.


\section*{Limitations}

Our work has a few limitations that should be acknowledged. 
First, we only used a single dataset for testing our proposed method. However, we made sure to use the largest publicly available dataset, MIMIC-CXR, which is about 50 times larger than IU-Xray \cite{demner2016preparing} the other widely-used dataset for radiology report generation. We focused on maximizing comparability to past and future work by employing the most complete set of evaluation metrics and by using the original train-validation-test split, unlike many other papers.
Secondly, while we obtain state-of-the-art performance on a variety of metrics by introducing our curriculum learning method, there is still room for research on the compatibility between this technique and the setup using an additional classification head. This would enable future approaches to benefit from both techniques, as they both led to impressive performance boost, individually.
Thirdly, while our work focused on simplicity, both in terms of the architecture used and on the introduction of the NLP-inspired curriculum learning method, based strictly on the report length, we acknowledge the fact that a different curriculum learning method designed specifically for medical imaging could improve the results even further. 
Finally, even though our results are state-of-the-art in many aspects, they are yet to be on par with radiologists performance in order for our approach to be used in real medical settings.

\bibliography{main}

\appendix

\clearpage
\appendix
\section{Appendix}
\label{sec:appendix}

\setcounter{secnumdepth}{2}
\setcounter{table}{0}
\setcounter{figure}{0}
\renewcommand{\thetable}{A\arabic{table}}
\renewcommand{\thefigure}{\thesection.\arabic{figure}}

\subsection{Qualitative Analysis of the Generated Reports}
\label{appendix:qualitative_analysis}

\begin{table*}[!htbp]
\centering
\small
\begin{tabular}{c|c|ccccccc}
\toprule
\textsc{Model} & \textsc{Report} & \textsc{BL1-4} & \textsc{RG-L} & \textsc{M}  \\ \hline
        \textsc{Context} & \begin{minipage}[t]{0.65\textwidth}
        {\_ year old male with history of metastatic melanoma, now with recurrent seizures and lethargy, comes here to evaluate for pneumonia.} \end{minipage} \\
        \midrule
        \textsc{Target} & \begin{minipage}[t]{0.65\textwidth}
        {impression : no acute cardiopulmonary process. findings : frontal and lateral radiographs of the chest redemonstrate a round calcified pulmonary nodule in the posterior right lung base, unchanged from multiple priors and consistent with prior granulomatous disease. a known enlarged right hilar lymph node seen on ct of \_ likely accounts for the increased opacity at the right hilum. a known right mediastinal lymph node conglomerate accounts for the fullness at the right paratracheal region. no pleural effusion, pneumothorax or focal consolidation is present. the patient is status post median sternotomy and cabg with wires intact. the cardiac silhouette is normal in size. the mediastinal and hilar contours are unchanged from the preceding radiograph.} \end{minipage} \\ 
        \midrule
        \textsc{Generated} & \begin{minipage}[t]{0.65\textwidth}
        {impression : no acute cardiopulmonary process. findings : frontal and lateral radiographs of the chest show no focal consolidation, pleural effusion or pneumothorax. the cardiomediastinal and hilar contours are normal. the patient is status post median sternotomy with intact wires.} \end{minipage} & \begin{minipage}[t]{0.05\textwidth}{0.165 0.135 0.117 0.106}\end{minipage} & $0.349$ & $0.330$ \\
        \bottomrule
        \textsc{Context} & \begin{minipage}[t]{0.65\textwidth}{post procedure, to assess for pneumothorax.}\end{minipage} \\
        \midrule
        \textsc{Target} & \begin{minipage}[t]{0.65\textwidth}{impression :  findings : in comparison with the study of \_, there is no evidence of pneumothorax. continued low lung volumes with substantial mass in the right paratracheal region.}\end{minipage} \\ 
        \midrule
        \textsc{Generated} & \begin{minipage}[t]{0.65\textwidth}{impression : findings : in comparison with the study of \_, there is little overall change. again there are low lung volumes with mild atelectatic changes at the bases. no evidence of acute focal pneumonia or vascular congestion.}\end{minipage} & \begin{minipage}[t]{0.05\textwidth}{0.571 0.501 0.455 0.412}\end{minipage} & $0.492$ & $0.701$  \\ 
        \bottomrule
        \textsc{Context} & \begin{minipage}[t]{0.65\textwidth}{picc.}\end{minipage} \\ 
        \midrule
        \textsc{Target} & \begin{minipage}[t]{0.65\textwidth}{impression : \_ left picc tip appears to terminate in the distal left brachiocephalic vein. \_ mild pulmonary vascular congestion. \_ interval improvement in aeration of the lung bases with residual streaky opacity likely reflective of atelectasis. interval resolution of the left pleural effusion. findings : left picc tip is seen terminating in the region of the distal left brachiocephalic vein. tracheostomy tube is in unchanged standard position. the heart is moderately enlarged. marked calcification of the aortic knob is again present. mild pulmonary vascular congestion is similar. bibasilar streaky airspace opacities are minimally improved. previously noted left pleural effusion appears to have resolved. no pneumothorax is identified. percutaneous gastrostomy tube is seen in the left upper quadrant.}\end{minipage} \\ 
        \midrule
        \textsc{Generated} & \begin{minipage}[t]{0.65\textwidth}{impression : findings : there has been interval removal of a right sided picc. the right sided picc line has been removed. tracheostomy tube and central venous catheter are again seen. there is a moderate cardiomegaly. there is hazy opacification of the right lung base which is stable. there is no overt pulmonary edema.} \end{minipage} & \begin{minipage}[t]{0.05\textwidth}{0.148 0.068 0.029 0.000}\end{minipage} & $0.170$ & $0.162$  \\
\bottomrule
\end{tabular}
\caption{Clinical accuracy metrics per pathology for our best model. All the predictions were generated using the \textsc{GIT-CXR (MV+C+CL)} model.}
\label{labelan}
\end{table*}

A few examples of generated reports, together with the corresponding context, target and evaluation metrics, are presented in Table \ref{labelan}. The examples were chosen to be representative for diverse scenarios (e.g. low, medium and high Meteor). Their analysis is made in the "Results and Evaluation" section.

\subsection{Experimental Setup}
\label{appendix:experimental_setup}

We train our models on a Linux system with 10 CPUs, 160GB RAM and an Nvidia H100 GPU with 80GB VRAM. 
Each model is trained for up to $N_e = 30$ epochs, using a patience of 7 epochs. The best checkpoint for each experiment was chosen on the validation set by using a weighted average of the NLG metrics used for evaluation, with weight 0.25 for METEOR and ROUGE-L and weight 0.125 for each of the BLEU scores. We call this score 

\vspace{2mm}
AVG\_NLG = $\frac{M}{4} + \frac{R}{4} +  \frac{B1+B2+B3+B4}{8}$.
\vspace{2mm}

We defined it so the selected checkpoint for each experiment will have competitive results all across the board.

We used the AdamW optimizer with a learning rate of $5\times10^{-5}$ and a batch size of 32 for all our experiments. The optimizer is similar to the one used for GIT \cite{wang2022git} pre-training, while the batch size 32 is the maximum power of 2 that would fit on the GPU for our biggest model. The learning rate was found by tuning the model GIT-CXR (SV+C) using the set \{ $10^{-5}$,  $5\times10^{-5}$\} for the AVG\_NLG score on the validation set. Then, it was used unchanged for all the models.

The images were resized to $224\times224$, similar to the GIT pre-training. The maximum length of the target report together with the context is set to 192 tokens and maximum length of context is 45 tokens out of the total 192. The 192 tokens has been chosen so that it would cover at least 94\% of the reports without requiring trimming.

In the case of the GIT-CXR-CLS, the weight of the classification loss is 0.1; this was chosen by tuning the model GIT-CXR-CLS (SV+C) using the set \{0.1, 1\} for the validation AVG\_NLG score.

In the case of the multi-view approach the number of views is set to 2, as less than 10\% of the studies contain 3 or more images.

For the curriculum learning approach we used the number of bins $b=10$ and at each epoch we sample $f=25\%$ of the dataset; because of this, we train the model for $\frac{1}{f}N_e=120$ epochs and validate every $\frac{1}{f}=4$ epochs, so the total number of samples passed through the model would be the same as for the models without curriculum learning. The fraction $f$ was found by tuning the model GIT-CXR (SV+C+CL) using the set \{0.1, 0.25\} for the AVG\_NLG score on the validation set.

For all experiments, we used the GIT model (the base variant GITb) pre-trained on the  MSRVTT-QA \cite{xu2017video} dataset, as this version of the model was adapted for video tasks, so it also contains pre-trained weights for the temporal embeddings required by the multi-view approach. 

In order to attain statistically significant results,
we ran each experiment 3 times and report the average of the results. The training time for each experiment varied between 24h for the base variant and up to 46h with the introduction of multi-view and multi-task approaches, for a total computational budget of roughly 1000 GPU hours for all the experiments combined.
We emphasize that all improvements of our curriculum learning based methods over the corresponding base models are statistically significant, according to a t-test with $p < 0.01$.

We will make our code publicly available.

\section{Dataset}
\label{appendix:dataset}

The MIMIC Chest X-ray Database v2.0.0 \cite{johnson2019mimicorig}, is the largest publicly available dataset containing chest X-ray images with their corresponding, free-text, clinical reports. The dataset contains a total of 377,110 DICOM format radiography images that correspond to 227,835 studies of 64,588 patients. This dataset is the basis of the MIMIC-CXR-JPG dataset, or MIMIC Chest X-ray JPG Database v2.0.0 \cite{johnson2019mimic}, that is entirely derived from MIMIC-CXR. Additionally processed, it provides JPG conversion of the original DICOM images and 14 pathologies (labels) for the reports using the CheXpert labeler~ \cite{irvin2019chexpert}. All of the 14 pathologies have 4 possible classes each (\textit{'Positive'}, \textit{'Negative'}, \textit{'Uncertain'} and \textit{'Missing'}). The MIMIC-CXR-JPG dataset offers the standard reference splits we also adopt in our study. Some studies have been manually reviewed by experts. The test set contains all the studies of patients who had at least one report labelled in the manual review. The validation set contains a random set of 500 patients and all of their associated studies. Finally, all the remaining studies are made available in the training set, resulting in a 222,758 - 1,808 - 3,269 distribution of train-validation-test studies. 

\paragraph{Studies} One patient can have one or more studies and one study can have one or more chest X-ray images and \textbf{exactly one} report. One report can contain one or more sections - free-text details about the patient's condition - such as \textit{'comparison'}, \textit{'clinical history'}, \textit{'indication'}, \textit{'reasons for examination'}, \textit{'impression'} and \textit{'findings'}. An example of a study can be seen in Figure \ref{fig:study}, where Figure \ref{fig:gen_report_ex} is the free-text report and the corresponding Figures \ref{fig:ap}, \ref{fig:pa}, \ref{fig:lat} are the chest X-ray images taken from different views of the patient.

\paragraph{Generation of Reports} Previous works use different sections of the reports. For example,  \cite{endo2021retrieval} uses both the \textit{'findings'} and the \textit{'impression'}, whereas \cite{nguyen2021automated}, \cite{lovelace2020learning} and \cite{miura2020improving} use only the \textit{'findings'} field. Considering the distribution of the reports in regards to the sections they contain, out of the 227,835 studies, 189,561 (83.2\%) reports contain an \textit{'impression'} section, and 155,716 (68.4\%) reports contain a \textit{'findings'} section. This adds up to 95.4\% of the entire dataset. Therefore, we decided to use both the \textit{'findings'} and the \textit{'impression'}, discarding the studies that do not contain reports with at least one of them. Additionally, they incorporate the most relevant information out of all the sections. We also pre-process the text and eliminate most symbols and upper-case letters. 

\paragraph{Images} The MIMIC-CXR-JPG dataset consists of chest X-ray images that are taken from different views of the patient - the front (Figure \ref{fig:ap}, anterior-posterior (AP)), the back (Figure \ref{fig:pa}, posterior-anterior (PA)), the lateral (Figure \ref{fig:lat}, LATERAL), or more particularly, the left-lateral (LL) part of the patient. In our experiments, we compare a \textbf{single-view} approach (also investigated by \citet{endo2021retrieval} and \citet{wang2022automated}) and a \textbf{multi-view} approach (also seen in \citet{nguyen2021automated} and \citet{miura2020improving}) of handling the situation involving multiple scans per study. 

\paragraph{Labels}  
For the labelling of our generated reports, we have chosen to use the CheXbert labeler by \citet{smit2020CheXbert}, a radiology report labelling method based on a biomedically pre-trained BERT \cite{devlin2018bert} that has near radiologist performance in labeling medical conditions and is 5.5\% more accurate than CheXpert. Related works that also use the CheXbert labeler are \cite{miura2020improving} and \cite{endo2021retrieval}, as opposed to works that use the CheXpert labeler \cite{chen2020generating, lovelace2020learning, nguyen2021automated}.

\section{Artifacts}
\label{appendix:artifacts}

Here we discuss the scientific artifacts used and created in this work.

Regarding data artifacts, as described in Section \ref{appendix:dataset}, we employ our method on the largest available dataset, namely MIMIC Chest X-ray Database v2.0.0 \cite{johnson2019mimicorig} and it's processed version MIMIC Chest X-ray JPG Database v2.0.0 \cite{johnson2019mimic}. Both are released under the licence \textit{PhysioNet Credentialed Health Data License 1.5.0}\footnote{https://physionet.org/content/mimic-cxr/view-license/2.1.0/} and data use agreement \textit{PhysioNet Credentialed Health Data Use Agreement 1.5.0}\footnote{https://physionet.org/content/mimic-cxr/view-dua/2.1.0/} after completing the required training \textit{CITI Data or Specimens Only Research}\footnote{https://physionet.org/content/mimic-cxr/view-required-training/2.1.0/}.

Regarding model artifacts, we used the base variant of the GIT model \cite{wang2022git}, released on Huggingface\footnote{https://huggingface.co/microsoft/git-base} under the \textit{MIT Licence}.

The evaluation for NLG metrics was done using the Huggingface library \footnote{https://huggingface.co/metrics}, while the clinical accuracy metrics were computed using the ChexBert labeler for classification and then the sklearn\footnote{https://scikit-learn.org/stable/modules/generated/sklearn.metrics.html} metrics P, R and F1 for evaluation. While we already provided a description in Section \ref{sec:evaluation_metrics}, the exact usage will also be provided in our code. 

Regarding the created artifacts, we will make our code, comprising all the approaches discussed in Section \ref{sec:method}, publicly available.

\newpage
\newpage
\newpage
\newpage
\newpage
\newpage
\newpage
\newpage
\newpage
\pagebreak

\end{document}